\documentclass[final]{cvpr}

\usepackage{times}
\usepackage{epsfig}
\usepackage{graphicx}
\usepackage{amsmath}
\usepackage{amssymb}
\usepackage{color,xcolor}
\usepackage{multirow}
\usepackage{subfigure}
\usepackage{setspace}

\usepackage[margin=0pt,font=small,labelfont=bf,labelsep=endash,tableposition=top]{caption}

\usepackage[pagebackref=true,breaklinks=true,colorlinks,bookmarks=false]{hyperref}

\begin{document}

\title{Graph Attention Tracking}

\author{Dongyan Guo$^{~\dagger}$, Yanyan Shao$^{~\dagger}$, Ying Cui$^{~\dagger}$, Zhenhua Wang$^{~\dagger}$, Liyan Zhang$^{~\ddagger}$,Chunhua Shen$^{~\sharp}$\\
	$^{\dagger}$~Zhejiang University of Technology, China\\$^{\ddagger}$~Nanjing University of Aeronautics and Astronautics, China\\$^{\sharp}$~University of Adelaide, North Terrace, Adelaide, SA 5005, Australia\\
	{\tt\small guodongyan@zjut.edu.cn}
}

\maketitle

\begin{abstract}

Siamese network based trackers formulate the visual tracking task as a similarity matching problem. Almost all popular Siamese trackers realize the similarity learning via convolutional feature cross-correlation between a target branch and a search branch. 
However, since the size of target feature region needs to be pre-fixed, these cross-correlation base methods suffer from either reserving much 
adverse 
background information or missing a great deal of foreground information. Moreover, the global matching between the target and search region also largely neglects the target structure and part-level information. 

In this paper, to solve the above issues, we propose a simple target-aware Siamese graph attention network for general object tracking.
We propose to establish part-to-part correspondence between the target and the search region with a complete bipartite graph, and apply the
graph attention  mechanism to propagate target information from
the 
template feature to
the
search feature. Further, instead of using the pre-fixed region cropping for template-feature-area selection, we investigate a target-aware area selection mechanism to fit the size and aspect ratio variations of different objects.
Experiments on challenging benchmarks including GOT-10k, UAV123, OTB-100 and LaSOT demonstrate that the proposed SiamGAT outperforms many state-of-the-art trackers and achieves leading performance. 
Code is available at: 
\url{https://git.io/SiamGAT}
\end{abstract}

\section{Introduction}

\begin{figure}[t]
	\centering 
		\includegraphics[width=0.48\textwidth]{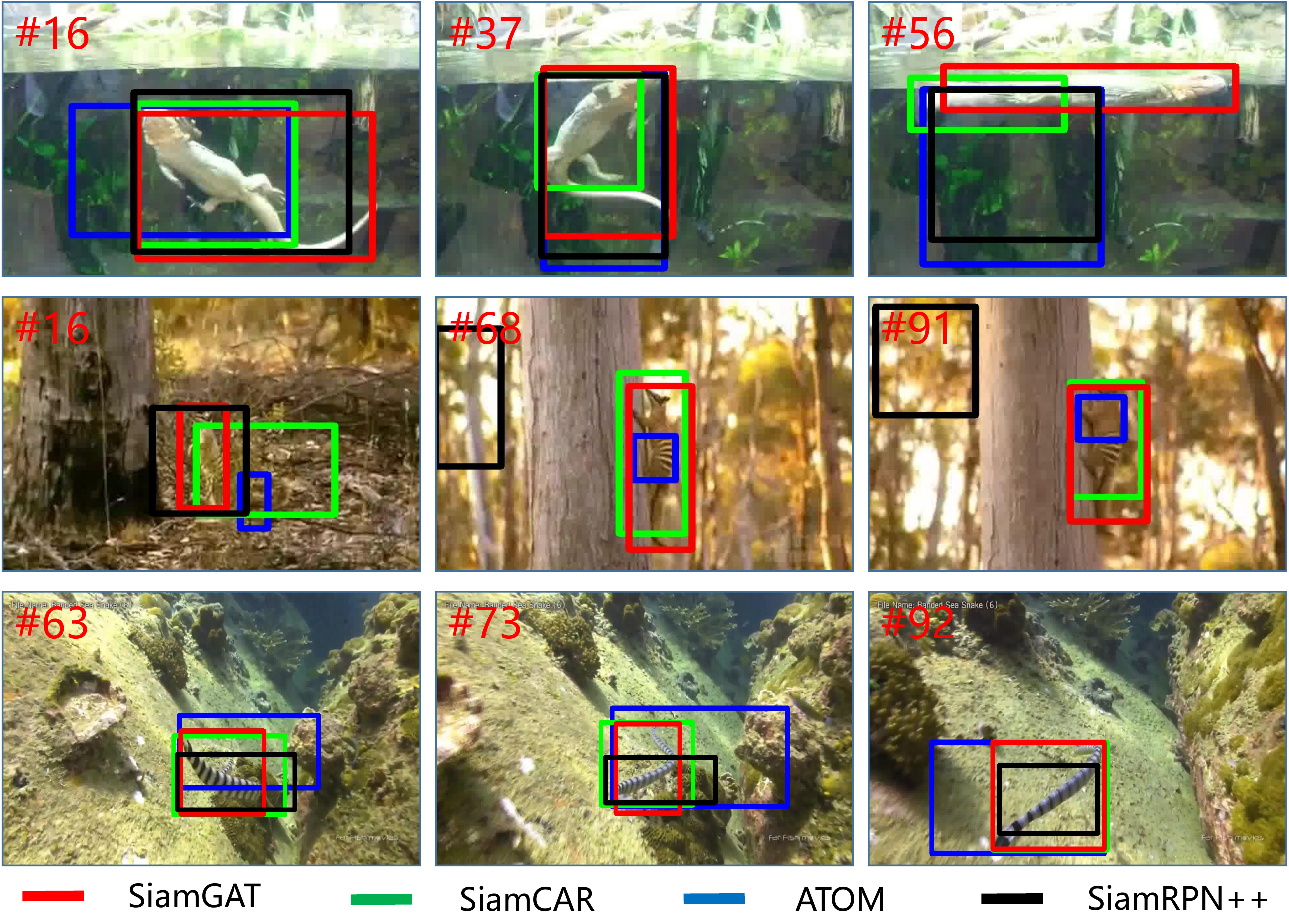}
	\caption{Comparisons of our SiamGAT with state-of-the-art trackers on three challenging sequences from GOT-10k.  Benefiting from the effective target information propagating, our SiamGAT successsfully handles the challenges 
	such as
	shape deformation, similar distractors and extreme aspect-ratio changes. Compared with the baseline SiamCAR (green), our SiamGAT (red) remarkably improves the tracking accuracy (zoom in for a better view). }
	\label{fig-comparision}
\end{figure}

\begin{figure}[t]
\centering 
		\includegraphics[scale=0.4]{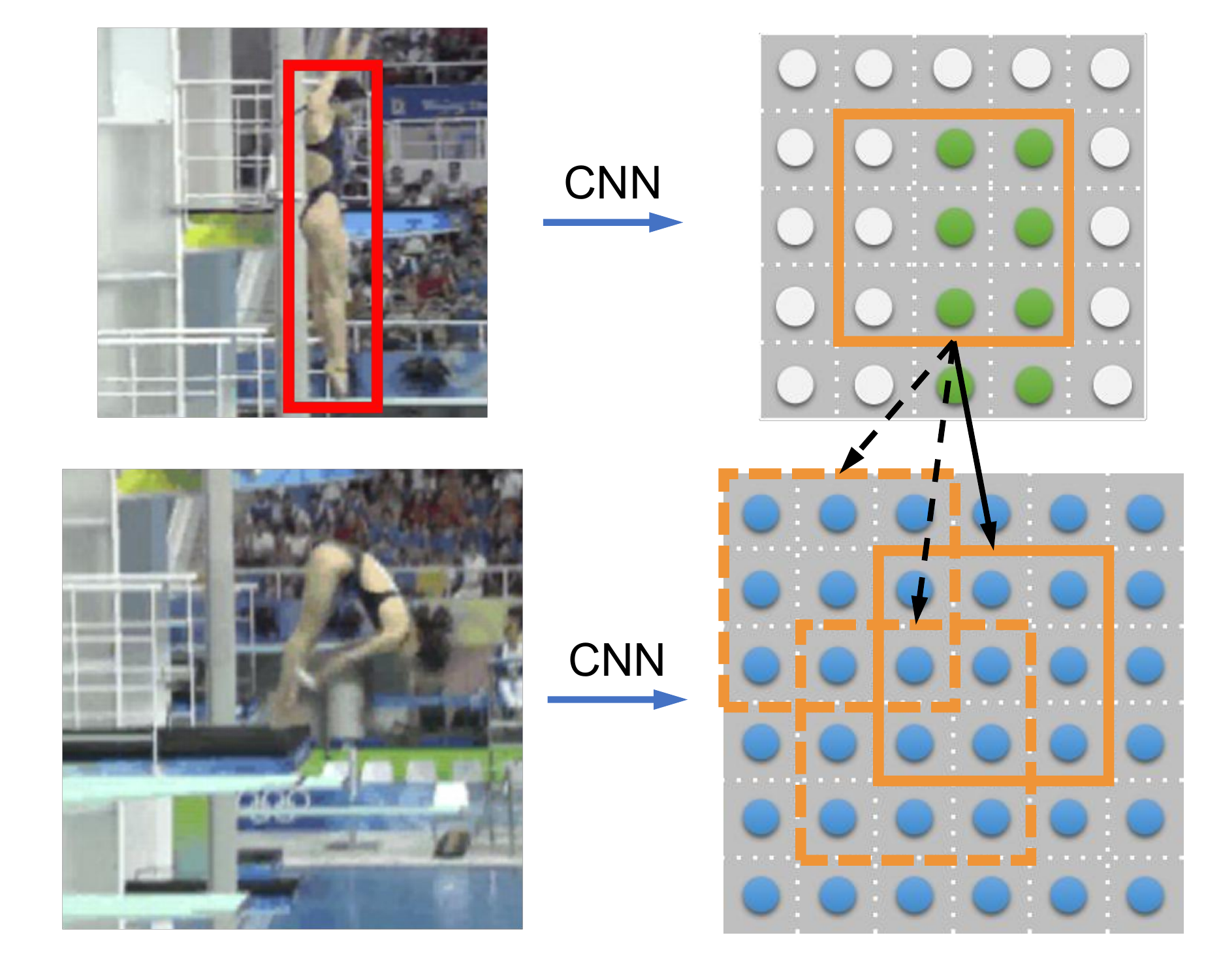}
	\caption{Illustration of 
	traditional cross-correlation based similarity learning methods. The target is marked by red boxes. The CNN features of the target, the background and the search region correspond to green, white and blue cycles respectively.  An important problem is that the template feature obtained by fixed-region cropping (
	labeled by the yellow box) may introduce 
	much
	background information or miss a great deal of foreground information, especially when the aspect ratio of the template target is 
	changed drastically.
	Moreover, during tracking, the target shape and pose are constantly changing, but the global matching 
	fails to consider the invariant part-level information and the transforming body shape.}
	\label{fig-Xcorr}
\end{figure}

General object tracking is a fundamental but challenging task in computer vision. In recent years, mainstream trackers focus on Siamese network based architectures \cite{SiamCAR, siamrpn++, SiamRPN,  SiamFC++}, which achieve state-of-the-art performance as well as 
a good balance between
tracking  accuracy and efficiency. 
These trackers first 
employ
a Siamese network for feature extraction. Then they develop a tracking-head network for object information decoding from one or more similarity maps (or so-called response maps) obtained by information embedding between the template-branch and the search-branch.
How to embed the information of the two branches to obtain informative response maps is a key issue, since information passed from the template to the search region is critical to the accurate localization of the object. 
Almost all current state-of-the-art Siamese trackers like SiamRPN  \cite{SiamRPN}, SiamRPN++ \cite{siamrpn++}, SiamFC++ \cite{SiamFC++} and SiamCAR \cite{SiamCAR} utilize a cross-correlation based layer for information embedding, which takes convolution on deep features as the basic operation.
Despite their great success,
some important drawbacks 
exist
with such cross-correlation based trackers: 1) The size of convolution kernel is pre-fixed. As shown in Figure~\ref{fig-Xcorr}, a common processing is cropping the central $m \times m$ region on the template feature map to generate the target feature, which is treated as the convolution kernel. However, when solving tracking tasks with different object scales or aspect ratios, this pre-fixed feature region may suffer from either reserving lots of background information or missing a great deal of foreground information, which consequently leads to inaccurate information embedding.
2) The target feature is treated as a whole for similarity computation with the search region. However, during tracking the target often yields large rotation, pose variation and heavy occlusions, and  performing such a global matching with variable target is not robust.
3) Because of 2), the information embedding between the template and search region is 
a global
information propagating process, in which the information transmitted from the template to the search region is limited and the information compression is excessive. Our key observation is that the information embedding should be performed by learning the part-level relations (instead of global matching), 
as 
part features tend to be invariant against shape and pose variations, thus being more robust.

Aiming at solving these issues, we leverage graph attention networks \cite{velickovic2018graph, segment} to design an \emph{part-to-part} information embedding network for object tracking.
We demonstrate that the information embedding between template and search region can be modeled with a complete bipartite graph, which encodes the relations between template nodes and search nodes by applying a graph attention mechanism \cite{velickovic2018graph}. With learned attentive scores, each search node can effectively aggregate target information from the template. All search nodes then yield a response map with 
rich
information for the subsequent decoding task.
With such designs, we propose a graph attention module (GAM) to realize \emph{part-to-part} information propagating instead of 
global
information propagating between the template and search region. Instead of using the whole template as a convolution kernel, this \emph{part-to-part} similarity matching can greatly alleviate the effect of shape-and-pose variations of targets. 
Further, instead of using the pre-fixed region cropping, we investigate a target-aware template computing mechanism to fit the size and aspect-ratio variations of different objects. 
With the introduced GAM and target-aware template computing techniques, we present a
novel
tracking framework, 
termed 
Siamese Graph Attention Tracking (SiamGAT) network,
for general object tracking. 

Since this work mainly argues that an effective information embedding algorithm can enhance the performance of 
the
tracking head,  the proposed SiamGAT simply consists of three essential blocks, without using any feature fusion, data enhancement or other strategies to enhance the performance. 
We evaluate our SiamGAT on several challenge benchmarks, including GOT-10k \cite{got10k}, OTB-100 \cite{wu2013online}, UAV123 \cite{uav123} and LaSOT \cite{lasot}. Without bells and whistles, the proposed tracker achieves leading performance
compared 
with state-of-the-art trackers. Our main contributions are 
as follows. 
\begin{itemize}
\itemsep 0pt
	\item
	
	We propose a graph attention module (GAM) to realize \emph{part-to-part} matching for information embedding. Compared with the traditional cross-correlation based approaches, the proposed GAM can greatly eliminate their drawbacks and effectively pass target information from template to search region.

    \item  We propose a target-aware Siamese Graph Attention Tracking (SiamGAT) network with GAM for general object tracking. The framework is simple yet effective. Compared with previous works using pre-fixed global feature matching, the proposed model is adaptive to the size and aspect-ratio variations of different objects.
  
    \item  Experiments on multiple challenging benchmarks including GOT-10k, UAV123, OTB-100 and LaSOT demonstrate that the proposed SiamGAT outperforms many state-of-the-art trackers and achieves leading performance.
    
\end{itemize}

\section{Related Work}
In recent years, Siamese based trackers have drawn great attention for their superior performance. 
The main structure of these trackers can be summarized as three parts: a Siamese network for feature extraction of the template and search region, a similarity matching module for information embedding of the two Siamese branches, and a tracking head for feature decoding from the similarity maps. Many researchers devote to optimizing the Siamese model for better feature representation, or designing new tracking head for more effective bounding box regression. However, few work has been done on information embedding.

The pioneering method SiamFC \cite{SiamFC} constructs a Siamese network model for feature extraction and utilizes a cross-correlation layer (Xcorr) to embed the two branches. It takes the template features as kernels to directly perform convolution operation on the search region and obtains a single channel response map. In essence, the correlation here can be regarded as a similarity calculation between the template and the search region, and the obtained response map is a similarity map for target location prediction. Following this similarity-learning work, many researchers try to enhance the Siamese model for feature representation but still leverage the cross-correlation for information embedding \cite{DSiam,saSiam,rasnet,gct}. DSiam \cite{DSiam} adds online learning modules to address the target appearance variation and background suppression transformation to improve feature representation. It focuses on enhancing the model updating ability, while the location of object is still computed based on the single channel response map. SA-Siam \cite{saSiam} utilizes a twofold Siamese network to train a semantic branch and an appearance branch. Each branch is a similarity-learning Siamese network, trained separately but combined at the testing time to complement each other. RASNet \cite{rasnet} introduces the spatial attention and channel attention mechanisms to enhance the discriminative capacity of the deep model. GCT \cite{gct} adopts a spatial-temporal graph convolutional network for target modeling. Since multiple scales are searched during test to handle the scale-variation of objects, these Siamese trackers  are time-consuming. 

Leveraging the region proposal network (RPN) \cite{ren2017faster} (proposed for object detection), Li \etal \cite{SiamRPN} propose the Siamese region proposal network SiamRPN.  They add two branches for region proposal at the end of the Siamese feature extraction network: one classification branch for background-foreground classification of anchors, and one regression branch for proposal refinement. To embed the information of anchors, SiamRPN \cite{SiamRPN} conducts an up-channel cross-correlation-layer (Up-Xcorr) by cascading multiple independent cross-correlation layers to output multi-channel response maps. Based on SiamRPN \cite{SiamRPN}, DaSiamRPN \cite{DaSiamRPN} designs a distractor-aware module to perform incremental learning and obtains much more discriminative features against semantic distractors. To tackle data imbalance, C-RPN \cite{CSiam} proposes to cascade a sequence of RPNs from deep high-level to shallow low-level layers in a Siamese network. Easy negative anchors can be filtered out in earlier cascade stage and hard samples are preserved across stages. Both SiamRPN++ \cite{siamrpn++} and SiamDW \cite{SiamDW}  investigate to deepen neural networks to improve the tracking performance. These RPN based trackers have achieved great success on performance as well as discarding traditional multi-scale tests. The chief drawback is that they are sensitive to hyper-parameters associated with anchors. 

Apart from deepening the Siamese network, SiamRPN++ \cite{siamrpn++} also presents a depth-wise cross-correlation layer (DW-Xcorr) to embed information of the target template and the search region branches. Specifically, it performs a channel-by-channel correlation operation with the feature maps of the two branches. By replacing the up-channel cross correlation with the depth-wise cross correlation, imbalance of parameter distribution of the two branches is resolved, which makes the training procedure more stable and the information association more efficient for the prediction of  bounding box. Later works in this vein devote to eliminate the negative effects of anchors. A number of  anchor-free trackers, such as SiamFC++ \cite{SiamFC++}, SiamCAR \cite{SiamCAR}, SiamBAN \cite{SiamBAN} and Ocean \cite{Ocean} are proposed, which achieve state-of-the-art tracking performance. They share the general idea 
tackling 
the tracking task as a joint classification and regression problem, and take one or multiple heads to directly predict objectiveness and regress bounding boxes from  response maps in a per-pixel-prediction manner. Ocean \cite{Ocean} further applies an online-updating module to 
dynamically adapt the tracker. 
By discarding anchors and proposals, these anchor-free trackers extricate from the tedious hyper-parameter-tuning and the requirement of providing prior information (\textit{e.g.}, data scale and ratio distribution) for the dataset.

Liao \etal \cite{PGNet} observed that traditional cross-correlation operation brings 
much
background information, which 
may
overwhelm the target feature and results in sensitivity to similar distracters. To solve this issue, they propose a pixel-to-global matching method to suppress the interference of background. However, similar to cross-correlation, this PG-correlation still takes a fixed-scale cropped region as the template feature.

\section{Method}

\begin{figure*}
	\centering 
	\includegraphics[scale=0.46]{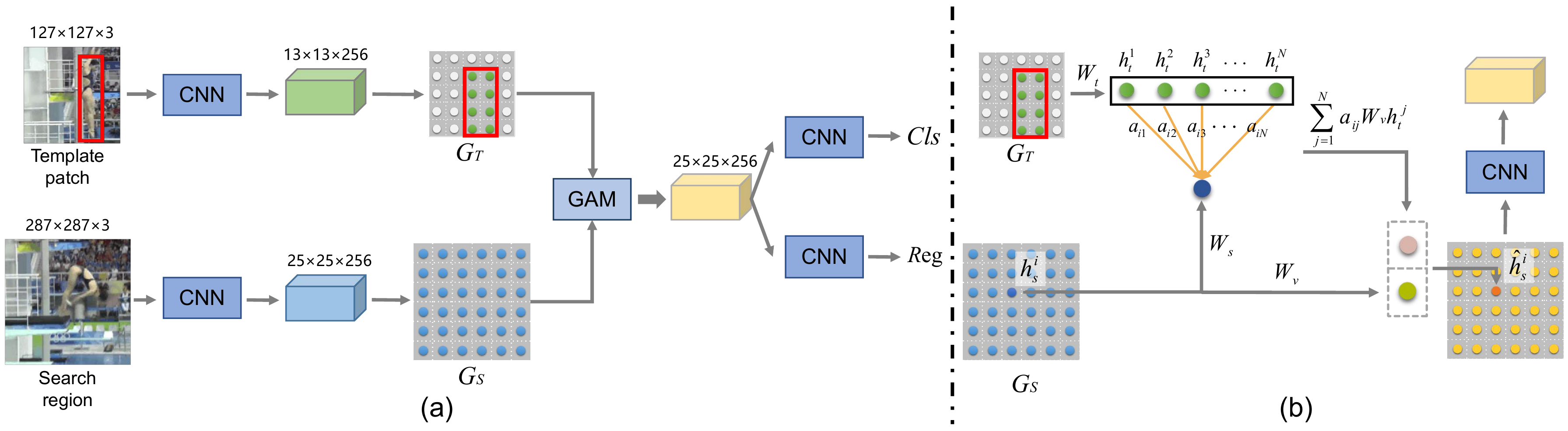}
	
	\caption{Overview of the proposed method. (a) The network architecture of SiamGAT. It consists of three primary blocks: a Siamese sub-network for feature extraction, a graph attention module for target information embedding, and a classification-regression sub-network for target localization. (b) Illustration of the proposed graph attention module. The representation of each search node is reconstructed by aggregating information from all neighboring target nodes with attention mechanism. Note that the number of target nodes is not fixed but varies with different target templates via a target-aware area selection mechanism.}
	\label{fig-framwork}
\end{figure*}
In this section, we 
present 
a detailed description for the proposed SiamGAT framework. The most important investigation of this work is that the performance of the Siamese trackers can be significantly improved with much effective information propagating from the target template to search region. In the following, we first introduce our Graph Attention Module which establishes the \emph{part-to-part} correspondence between the Siamese branches. 
Then
we present the target-aware graph attention tracker. An overview of our framework is illustrated 
in
Figure~\ref{fig-framwork}. 

\subsection{Graph Attention Information Embedding} \label{gam}
Existing correlation based information embedding methods \cite{SiamFC, SiamRPN, siamrpn++}  take the whole target feature as a unity to match with the search features. As this operation neglects the part-level correspondence between the target and the search regions, the matching is inaccurate under shape-and-pose variances of targets. Besides, this
global
matching manner may greatly compress the target information propagating to the search feature. In order to address these problems, we establish the \emph{part-to-part} correspondence between the target template and the search region with a complete bipartite graph.   

Given two images of a template patch $T$ and a search region $S$, we first employ a Siamese feature extraction network to obtain two feature maps $F_t$ and $F_s$. To generate a graph, we  consider each $1\times 1\times c$ grid of the feature map as a node (part), where $c$ represents the number of feature channels. Let $V_t$ be a node set including all nodes of $F_t$, and let $V_s$ be another node set of $F_s$. Inspired by the graph attention networks \cite{velickovic2018graph}, we 
use
a complete bipartite graph $G=(V,E)$ to model the part-level relations between the target and search region, where $V=V_s\cup V_t$ and $E=\{(u,v)|\forall u\in V_s, \forall v\in V_t\}$. We further define two sub-graphs of $G$ by $G_t=(V_t,\emptyset)$ and $G_s=(V_s,\emptyset)$. 

For each $(i,j)\in E$, 
let $e_{ij}$ denote the correlation score of node $i\in V_s$ and node $j\in V_t$:
\begin{equation}
    e_{ij}=f(\mathbf{h}_s^i,\mathbf{h}_t^j),
\end{equation}
where $\mathbf{h}_s^i \in \mathbb{R}^c$ and $\mathbf{h}_t^j \in \mathbb{R}^c$ are feature vectors of node $i$ and node $j$. Since the more similar is
a location in the search region  to the local features of the template, the more likely it is the foreground, and 
more target information should be passed to there. For this reason, we hope that the score $e_{ij}$ is  proportional to the similarity of the two node features. 
We can 
simply use
the inner product between features as the similarity measurement. In order to adaptively learn a better representation between the nodes, we first apply linear transformations to the node features and then take the inner product between transformed feature vectors to calculate the correlation score. Formally,
\begin{equation}
f(\mathbf{h}_s^i,\mathbf{h}_t^j)=(W_s \mathbf{h}_s^i)^T(W_t \mathbf{h}_t^j),
\end{equation}
where $W_s$ and $W_t$ are the linear transformation matrices.

In order to balance the amount of information sent to the search region, we normalize $e_{ij}$ with the softmax function:
\begin{equation}
a_{ij}=\frac{\exp(e_{ij})}{\sum_{k\in V_t} \exp(e_{ik})}.
\end{equation}
Intuitively, $a_{ij}$ measures how much attention the tracker should pay to part $i$, according to the viewpoint of part $j$.

Leveraging the attentions  that passed from all nodes in $G_t$ to the $i$-th node in $G_s$, we compute the aggregated representation for node $i$ with
\begin{equation}
\mathbf{v}_i=\sum_{j\in V_t}a_{ij}W_v \mathbf{h}_t^j,
\end{equation}
where $W_v$ is a matrix for linear transformation.

Finally, we can fuse the aggregated feature with the node feature $\mathbf{h}_s^i$ to obtain a more powerful feature representation empowered by target information:
\begin{equation}
\hat{\mathbf{h}}_s^i={\rm ReLU}\big(\mathbf{v}_i\|(W_v \mathbf{h}_s^i)\big),
\end{equation}
where $\|$ represents vector concatenation. 

We compute all $\hat{\mathbf{h}}_s^i~\forall i \in V_s$ in parallel, which yields a response map for subsequent task.

\subsection{Target-Aware Graph Attention Tracking}

We have presented the graph attention module (GAM) to realize the \emph{part-to-part} information propagating. Before achieving target-aware visual tracking, we need to tackle 
 another 
challenge. 
That is, how to 
produce
a variable template which adaptively fits different object scales and aspect-ratios.   

Traditional cross-correlation based methods simply crop the center region of template $F_t$ as the target feature to match with the search region $F_s$, which delivers much background information to the response map, especially when the template target is given in extreme aspect ratios. 
To address the problem, we investigate a target-aware template-feature-area selection mechanism under the supervision of  labeled bounding box $B_t$ in the template patch.  
By projecting $B_t$ onto the feature map $F_t$, we can 
attain
a region of interest $R_t$. Only the pixels in $R_t$ are taken as the template feature: 
\begin{equation}
\hat{F}_t=\bigg[ F_t(i,j,:)\bigg]_{(i,j) \in R_t}.
\end{equation}
Through this simple operation, the obtained feature map $\hat{F}_t$ is a tensor
of
dimensions $(w,h,c)$, where $w$ and $h$ correspond to the width and height of the template bounding box $B_t$, and $c$ is the number of channels of $F_t$.

Each element $\hat{F}_t(i,j,:)$ is considered as a node in the template subgraph $G_t$. Meanwhile, each element $F_s(m,n,:)$ is considered as a node in the search subgraph $G_s$. These two subgraphs serve as inputs to the Graph Attention Module for information embedding. As elements in $G_t$ are arranged in a grid pattern on the feature map $\hat{F}_t$, we can implement the linear transformations in Section~\ref{gam} with $1 \times 1$ convolutions. Then all correlation scores could be calculated by matrix multiplication, 
which is expected to greatly improve the efficiency.

In experiments, we observe that applying a batch normalization after each convolution can effectively improve the performance.
However, the dimensions $w$ and $h$ corresponding to different tracking objects cannot be pre-determined, thus we cannot directly apply the batch normalization operation with the scale variable $\hat{F}_t$. To solve the problem, we recompute $\hat{F}_t$
as follows:
\begin{equation}
\hat{F}_t(i,j,:)=\begin{cases}
F_t(i,j,:) & {\rm if~} (i,j) \in R_t, \\
0 & {\rm otherwise}.
\end{cases}
\end{equation}
Besides keeping the scale invariant, this target-aware idea renders the proposed method extendable to tasks which require non-rectangular ROIs (\textit{e.g.}, instance-segmentation in videos) .

Now we can construct our tracking network with the proposed GAM for effective information embedding. As shown in Figure~\ref{fig-framwork}(a), our SiamGAT simply consists of three blocks: a Siamese network for feature extraction, a tracking head for target bounding box prediction and a GAM block to bridge them.

Numerous works demonstrated that trackers can greatly benefit from better feature extraction method \cite{staple,siamrpn++}. By replacing the classical HOG features and color features with deep CNN features, tracking accuracy has seen significant improvement \cite{staple}. Later, deepening the backbone networks and fusing features of multiple layers  have further improved the tracking performance \cite{siamrpn++}. 
Since GoogLeNet \cite{googlenet-v3} is able to learn multi-scale feature representations with much fewer parameter and faster reasonging speed, here we adopt GoogLeNet as our backbone (an ablation is performed as well to study the performance of SiamGAT using different backbones).

Encouraged
by the success of anchor-free trackers, we leverage the classification-regression head network from SiamCAR \cite{SiamCAR} 
to be
the tracking head. It contains two branches: a classification branch predicting the category information for each location, and a regression branch computing the target bounding box at this location. The two branches share the same response map output by GAM.

\begin{table*}
	\setlength{\abovecaptionskip}{0.cm}
	\setlength{\belowcaptionskip}{-0.4cm}	
	\vspace{-10pt}
	\begin{center}
		\setlength{\tabcolsep}{5mm}
		\begin{spacing}{1.20}
			\begin{tabular}{c|cccccc}
			\hline
				Dataset& Backbone& Target-aware & Embedding Type & Success & Precision & FPS \\
				\hline
				\multirow{4}{*}{UAV123} & AlexNet & $\checkmark$  & GAM & 0.592 & 0.779 & 165 \\
				& GoogLeNet& $\checkmark$ & GAM & \textbf{0.646} & \textbf{0.843} & 70 \\
				& GoogLeNet& $\times$ & GAM & 0.626 & 0.822 &  71\\
				& GoogLeNet& $\times $ &	DW-Xcorr & 0.615 & 0.815 &  74\\
			\hline
			\end{tabular}
		\end{spacing}
	\end{center}
	\vspace{-0pt}
	\caption{Ablation study on UAV123. Target-aware represents whether the template feature area is pre-fixed or adptively selected with the object aspect ratio.}
	\label{tab-Ablation}
\end{table*}

\begin{figure*}[htbp]
	\centering
	\subfigure{
		\begin{minipage}[t]{0.50\linewidth}
			\centering
			\includegraphics[width=3.2in]{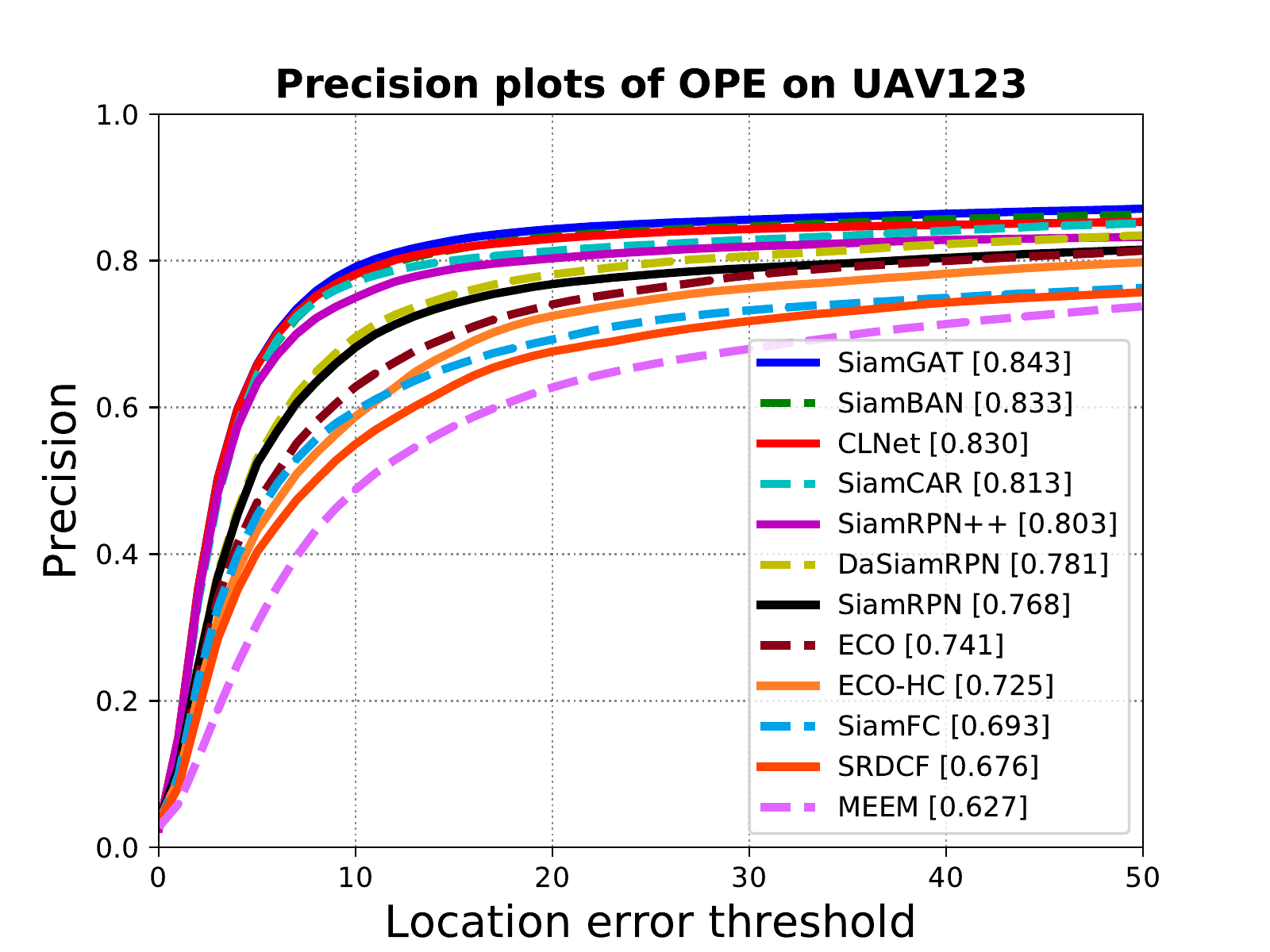}
		\end{minipage}%
	}%
	\subfigure{
		\begin{minipage}[t]{0.50\linewidth}
			\centering
			\includegraphics[width=3.2in]{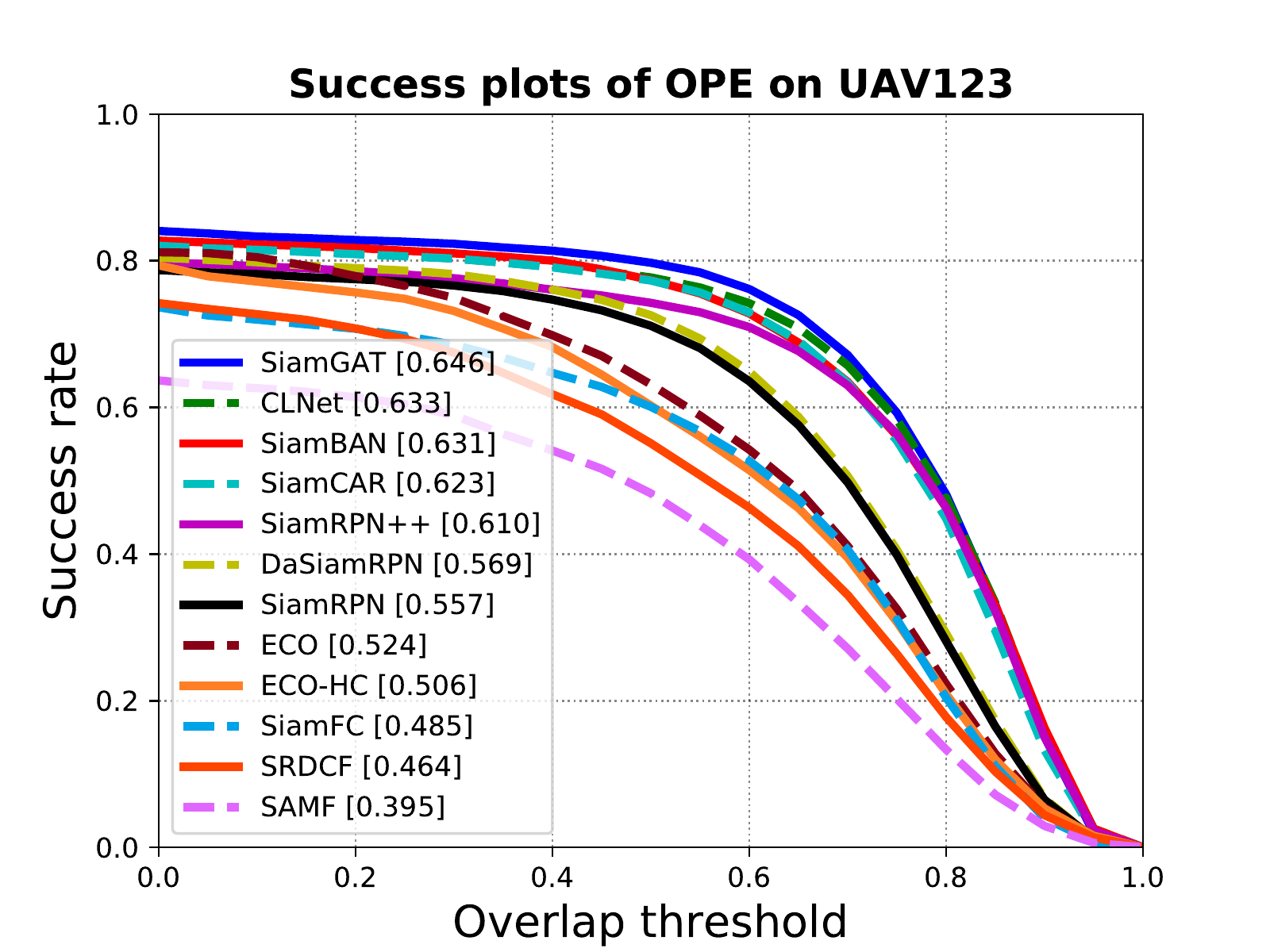}
		\end{minipage}
	}%
	\caption{Comparisions with state-of-the-art tracker on UAV123 \cite{uav123} in terms of precision plots of OPE and success plots of OPE.}
	\label{fig-uav123}
\end{figure*}

\section{Experiment}

\subsection{Implementation Details}
The proposed SiamGAT is implemented in
Pytorch on 4 RTX-2080Ti cards. Unless specified, the modified GoogLeNet( Inception v3) \cite{googlenet-v3} is adopted as the backbone network for feature extraction. 
The backbone is initialized with the weights that pretrained on ImageNet \cite{russakovsky2015imagenet}. The training batch size is set as 76 and totally 20 epochs are trained with stochastic gradient descent (SGD). We use a learning rate that linearly increased from 0.005 to 0.01 for the first 5 warmup epoches and then exponentially decayed to 0.0005 for the rest 15 epoches. For the  first 10 epoches, we freeze the parameters in the backbone to train the graph attention network and the head network. For the rest 10 epoches, we freeze stage 1 and 2 of GoogLeNet, and fine-tune stage 3 and 4. 

We adopt COCO \cite{lin2014microsoft}, ImageNet DET \cite{russakovsky2015imagenet}, ImageNet VID \cite{russakovsky2015imagenet}, YouTube-BB \cite{real2017youtube} and GOT-10k \cite{got10k} as the 
training set for experiments on OTB100 \cite{wu2013online} and UAV123 \cite{uav123}. 
Specifically, for experiments on GOT-10k \cite{got10k} and LaSOT \cite{lasot}, the model is respectively trained with only the specified training set provided by their official websites for fair comparison.
In both training and testing processes, we use pre-fixed scales with $127 \times 127$ pixels for the template patch and $287 \times 287$ pixels for search regions. During testing, only the object in the initial frame of a sequence is adopted as the template patch and fixed for the whole tracking period of this sequence. The search region in the current frame is adopted as the input of the search branch.

\subsection{Ablation Study}
\textbf{Backbone architecture.} 
We evaluate our network with both shallow and deep backbone architectures for visual tracking.
Table~\ref{tab-Ablation} shows the tracking performance with AlexNet and GoogLeNet as backbones.
Different backbones greatly affect the speed and performance of the tracker. By replacing AlexNet with GoogLeNet, the success is improved by $5.4\%$ from $59.2\%$ to $64.6\%$, the precision is increased by $6.4\%$ from $77.9\%$ to $84.3\%$. While the tracking speed decreases from 165 FPS to 70 FPS, which still meets the real-time requirement. 
It is worth pointing out that,
the SiamGAT using  AlexNet as the backbone also achieves a competitive performance while its precision and success are $1.1\%$ and $3.5\%$  higher than SiamRPN \cite{SiamRPN}, whose results are shown in Figure~\ref{fig-uav123}. 
Clearly, the proposed approach can achieve a trade-off between accuracy and efficiency with different backbones.

\textbf{Target-aware vs.\  pre-fixed template area selection.} 
To investigate the impact of template area selection, we train two  models with GAM on GoogLeNet. One is trained with the traditional fixed-region cropping target features, and another is trained with the target-aware selected features. As shown in Table~\ref{tab-Ablation}, the proposed target-aware feature area selecting mechanism brings $2.0\%$ and $2.1\%$ performance gains respectively on success and precision. The main reason is that the target-aware mechanism is able to effectively eliminate the background information and enhance the foreground representation, which helps to obtain more accurate target feature area than fixed-region cropping.

\textbf{Comparison with DW-Xcorr.}
To conduct a comparison with cross-correlation based methods, here we  replace the target-aware GAM with the popular DW-Xcorr layer \cite{siamrpn++}, which achieves the best performance among cross-correlation based methods. As shown in Table~\ref{tab-Ablation}, compared with DW-Xcorr, the GAM with target-aware mechanism brings $3.1\%$ and $2.8\%$ performance gains respectively on success and precision, while the GAM with pre-fixed region only brings $1.1\%$ and $0.7\%$ performance gains. The results further demonstrate that the pre-fixed region of target features has become 
the 
bottleneck for accurate target-information-passing. Benefiting from the GAM architecture, our method enables the target-aware region, which is adaptive to different aspect ratios of objects. 

\subsection{Evaluation on UAV123}
The 
UAV123 dataset contains a total of 123 video sequences and all sequences are fully annotated with upright bounding boxes. Objects in the dataset suffer from occlusions, fast motion, illumination and large scale variations, which
pose 
challenges to the trackers. 
A comparison with state-of-the-art trackers is shown in Figure~\ref{fig-uav123} in terms of the precision and success plots of OPE. Our tracker outperforms all other trackers for both metrics. Compared with the baseline SiamCAR, our tracker improves the performance 
by
$3.0\%$ in precision and $2.3\%$ in success.

\subsection{Evaluation on GOT-10k}
\begin{table}
	\begin{center}
		\begin{tabular}{r |ccc}
			\hline
			Tracker & AO & SR0.5 & SR0.75 \\
			\hline
			CFNet \cite{cfnet} & 29.3 & 26.5 & 8.7 \\
			MDNet \cite{mdnet} & 29.9 & 30.3 & 9.9 \\
			ECO \cite{ECO} & 31.6 & 30.9 & 11.1 \\
			CCOT \cite{CCOT} & 32.5 & 32.8 & 10.7 \\
			GOTURN \cite{GOTURN} & 34.7 & 37.5 & 12.4 \\
			SiamFC \cite{SiamFC} & 34.8 & 35.3 & 9.8 \\
			SiamRPN\_R18 \cite{SiamRPN} & 48.3 & 58.1 & 27.0 \\
			SPM \cite{spm} & 51.3 & 59.3 & 35.9 \\
			SiamRPN++ \cite{siamrpn++} & 51.7 & 61.5 & 32.9 \\
			ATOM \cite{atom} & 55.6 & 63.4 & 40.2 \\
			SiamCAR \cite{SiamCAR} & 57.9 & 67.7 & 43.7 \\
			SiamFC++ \cite{SiamFC++} & 59.5 & 69.5 & 47.9 \\
			D3S \cite{D3S} & 59.7 & 67.6 & 46.2 \\
			Ocean-offline \cite{Ocean} & 59.2 & 69.5 & 47.3 \\
			Ocean-online \cite{Ocean} & 61.1 & 72.1 & 47.3 \\

			SiamGAT (ours) & \textbf{62.7} & \textbf{74.3} & \textbf{48.8} \\
			\hline
		\end{tabular}
	\end{center}
	\caption{Evaluation on GOT-10k \cite{got10k} in terms of average overlap and success rate. Our SiamGAT 
	achieves best results.}
	\label{tab-GOT10k}
\end{table}

\begin{figure}
	\centering 
	\includegraphics[width=0.47\textwidth]{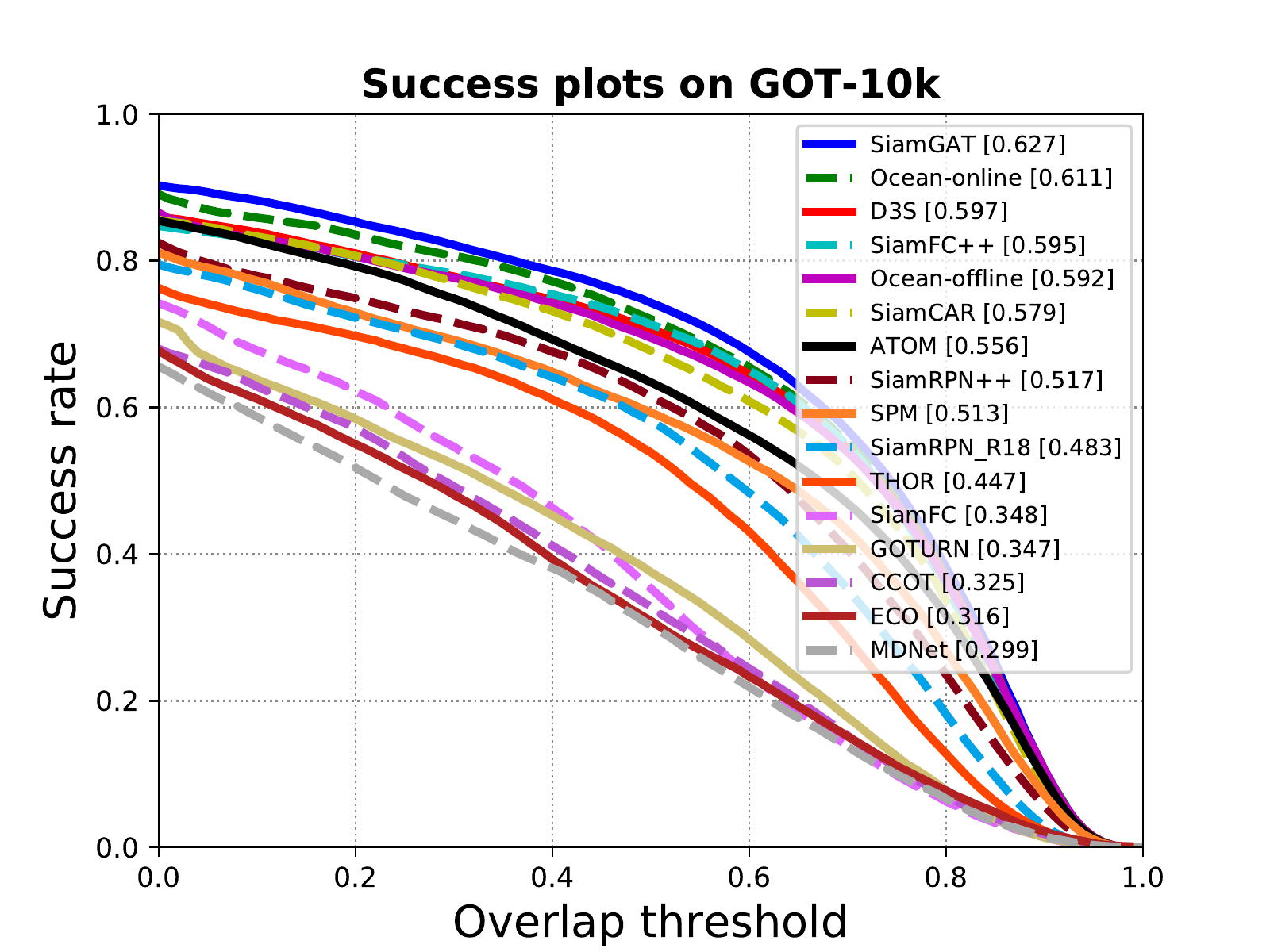}
	\caption{A comparison of our SiamGAT with state-of-the-art trackers in terms of success plots on GOT-10k \cite{got10k}.}
	\label{fig-GOT-10K}
\end{figure}

\begin{figure}[htbp]
	\centering

	\includegraphics[width=0.48\textwidth]{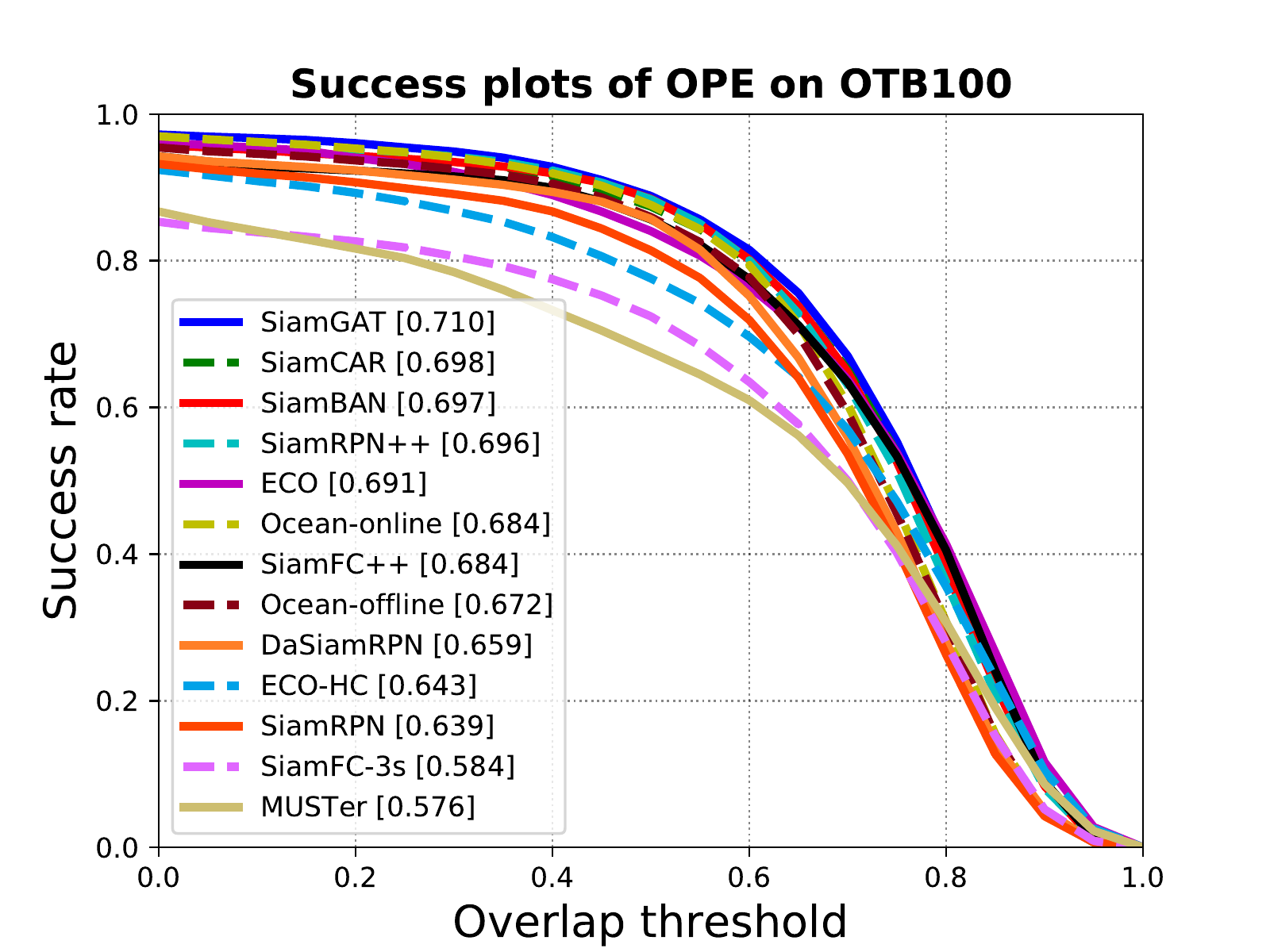}
	\caption{Comparision with state-of-the-art trackers on OTB-100 \cite{wu2015object} in terms of success plots of OPE.}
	
	\label{fig-otb2015}
\end{figure}

\begin{figure}[t]
	\begin{center}
		\includegraphics[scale=0.38]{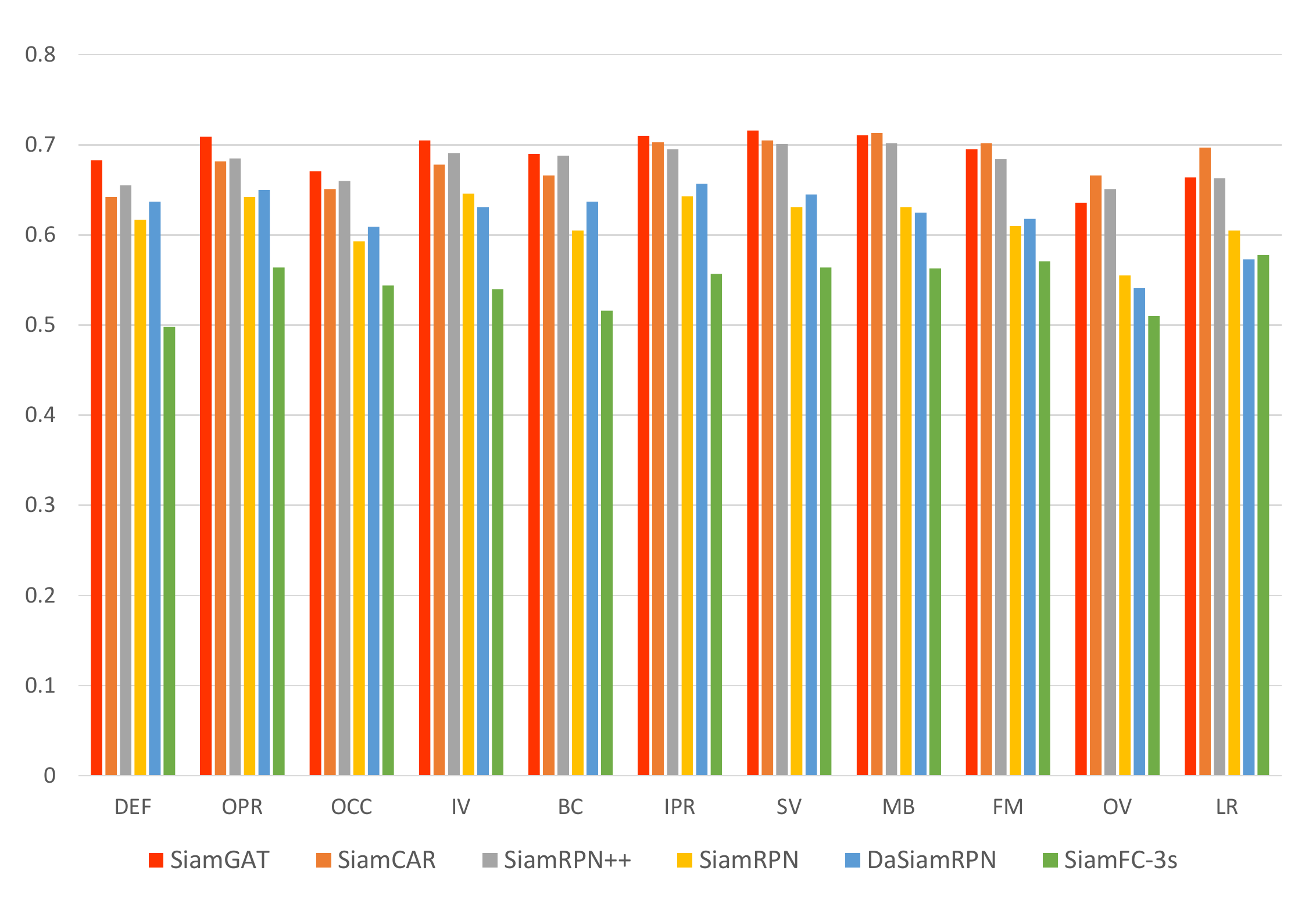}
	\end{center}
	\caption{Evaluation with each single attribute on OTB-100 \cite{wu2015object} in terms of success.}
	\label{fig-attributes}
\end{figure}

\begin{figure*}[htbp]
	\centering
	\subfigure{
		\begin{minipage}[t]{0.33\linewidth}
			\centering
			\includegraphics[width=2.2in]{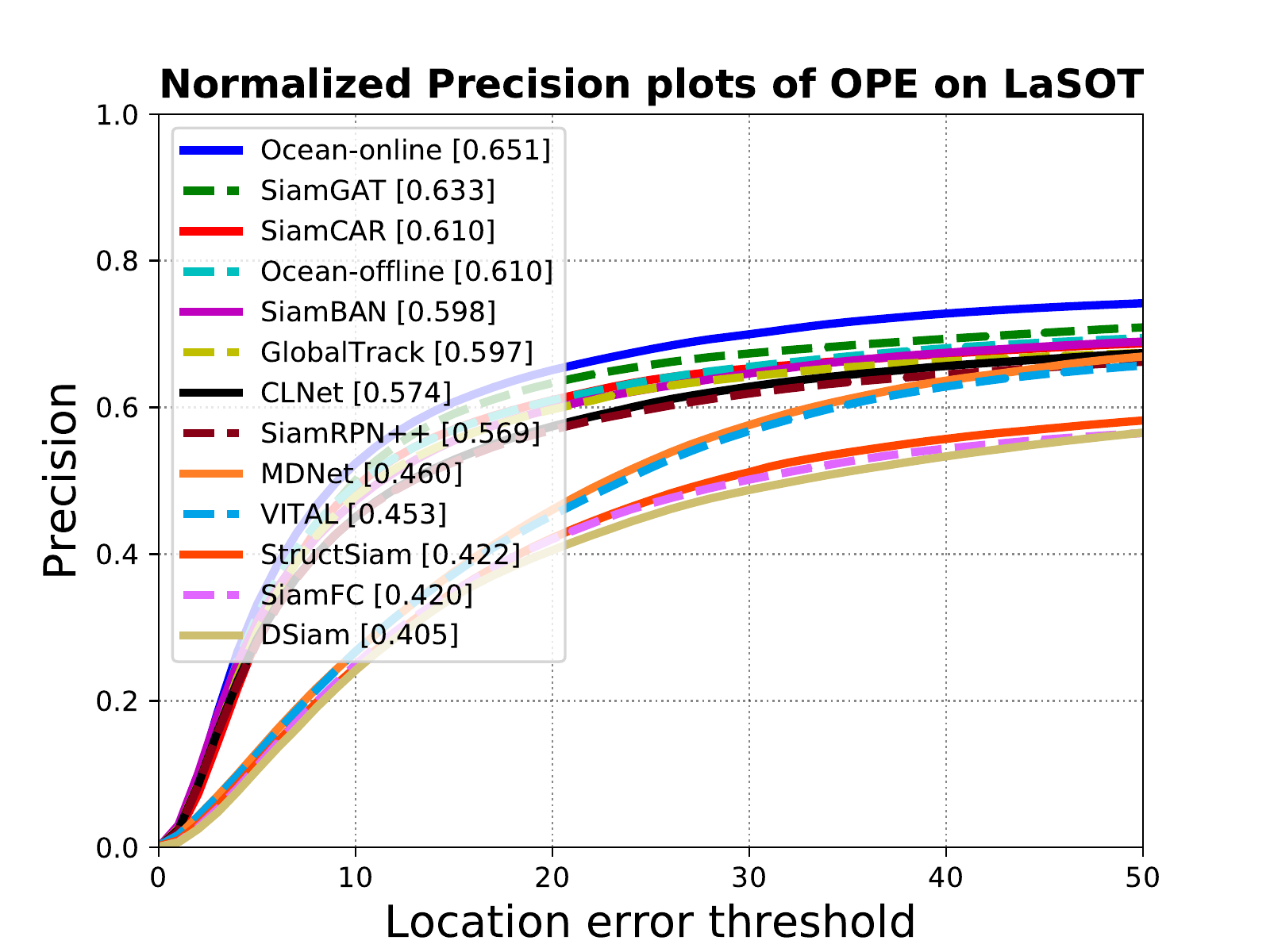}
		\end{minipage}%
	}%
	\subfigure{
		\begin{minipage}[t]{0.33\linewidth}
			\centering
			\includegraphics[width=2.2in]{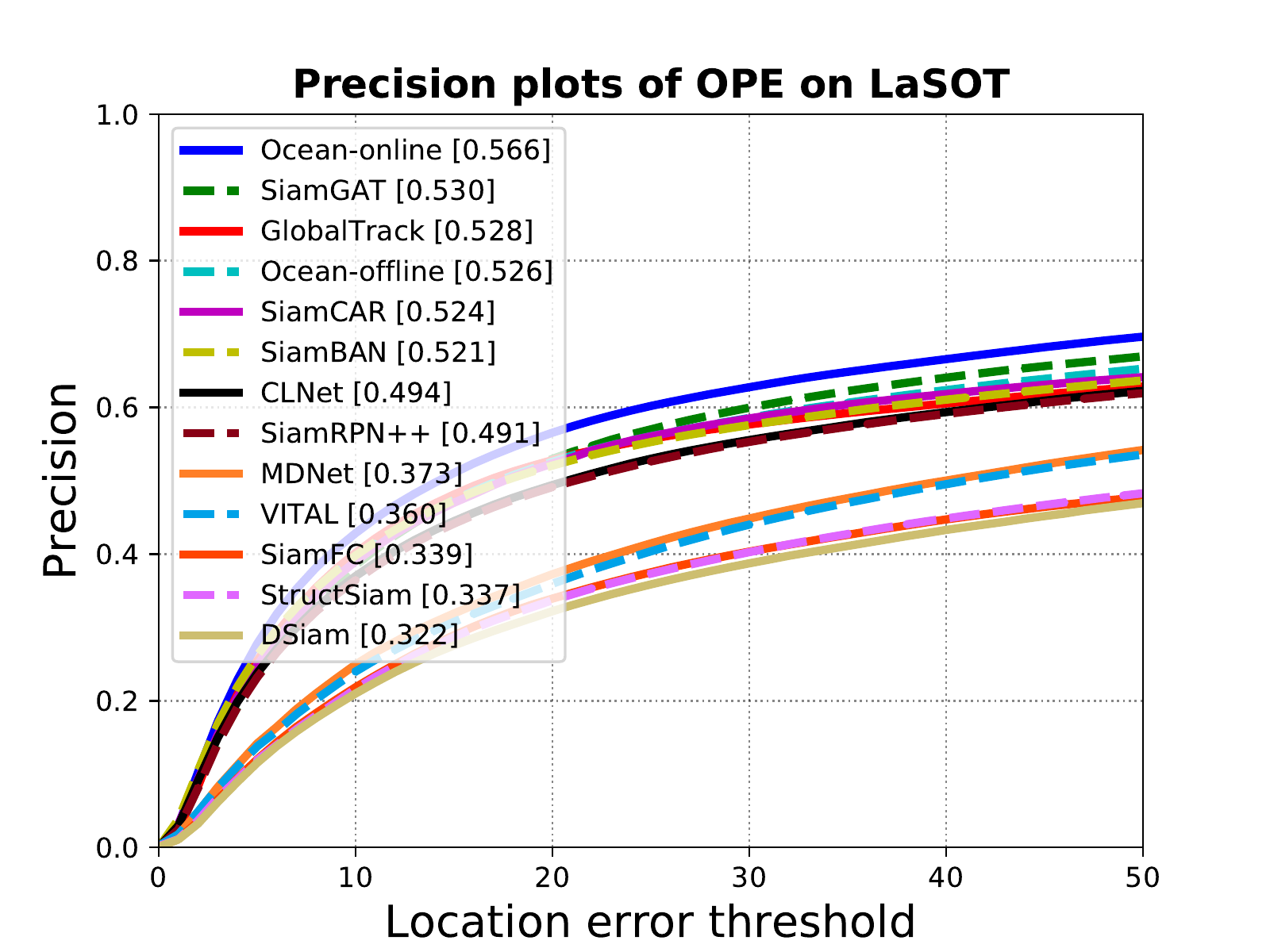}
		\end{minipage}
	}%
	\subfigure{
		\begin{minipage}[t]{0.33\linewidth}
			\centering
			\includegraphics[width=2.2in]{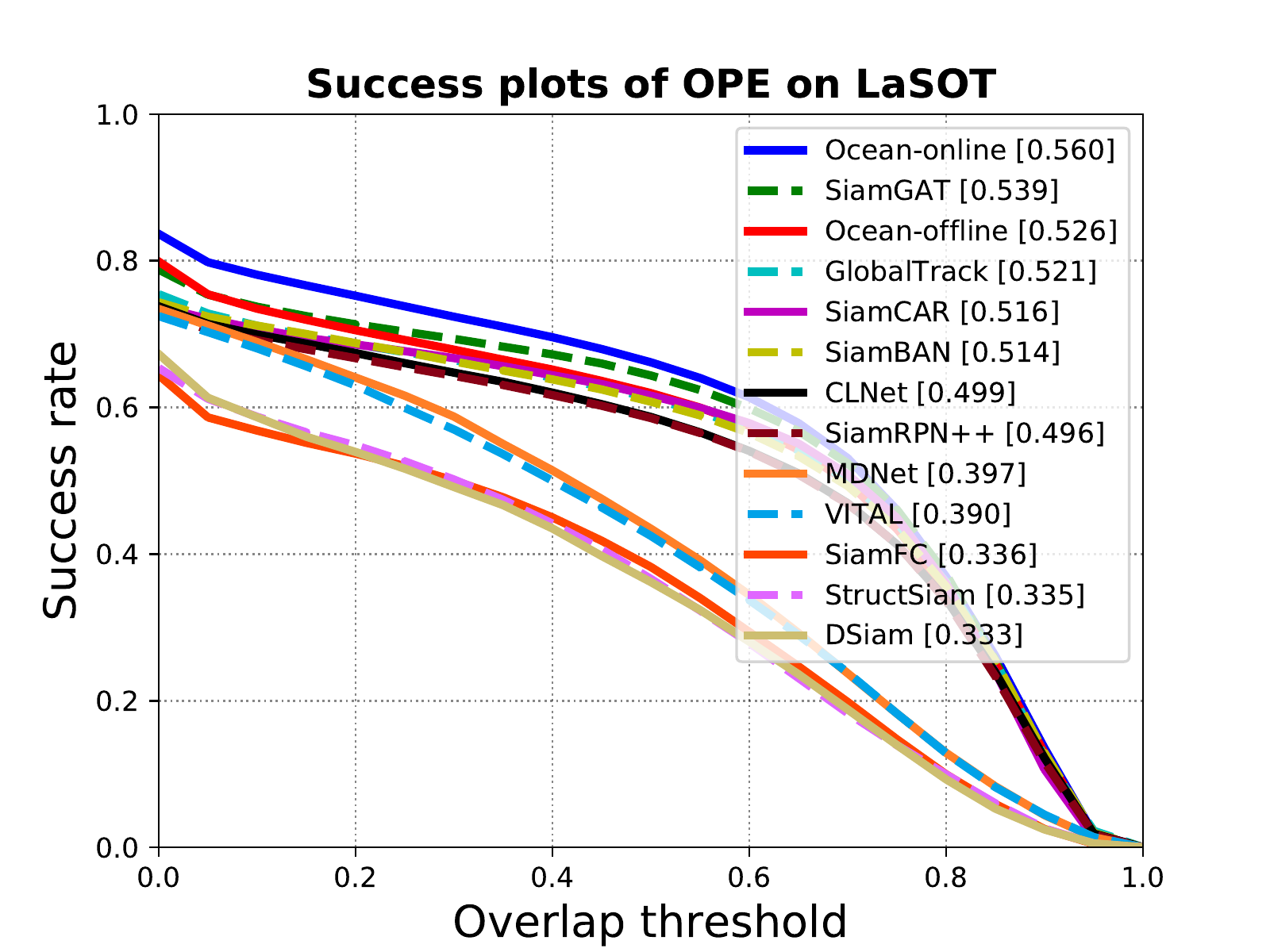}
		\end{minipage}
	}%
	\centering
	\caption{ Comparision with state-of-the-art trackers on LaSOT \cite{lasot} in terms of the normalized precision, precision and success plots of OPE.}
	\label{fig-lasot}
\end{figure*}
To evaluate the generalization of our tracker, we test it on the GOT-10k (Generic Object Tracking Benchmark) and compare it with state-of-the-art trackers. GOT-10k is a challenging large-scale dataset which contains more than 10,000 videos of moving objects in real-world.  It 
is also
challenging  in terms of zero-class-overlap between the provided training subset and testing subset. For fair comparison, we follow the the protocol of GOT-10k that only training our model with its training subset. 
We evaluate SiamGAT on GOT-10k and compare it with state-of-the-art trackers including SiamCAR \cite{SiamCAR}, Ocean \cite{Ocean}, SiamFC++ \cite{SiamFC++}, D3S \cite{D3S}, SiamRPN++ \cite{siamrpn++}, SPM \cite{spm}, SiamRPN \cite{SiamRPN} and other baselines. As shown in Table~\ref{tab-GOT10k}, the proposed SiamGAT performs best in term of all metrics. Compared with the baseline SiamCAR, our tracker improves by $4.8\%$, $6.6\%$ and $5.1\%$  respectively 
in terms of
$AO$, $SR_{0.5}$ and $SR_{0.75}$.  
Impressively, it even outperforms the online update tracker `Ocean' and improves the scores respectively by $1.6\%$, $2.2\%$ and $1.5\%$ with a much simple network architecture, which validates the generalization ability of our tracker on unseen classes. 
Figure~\ref{fig-GOT-10K} shows a comparison on success plots. Some qualitative results and comparisons are provided by Figure~\ref{fig-comparision}, which demonstrates that our SiamGAT is able to predict  more accurate bounding boxes of targets.

\subsection{Evaluation on OTB-100}

OTB-100 is one of the most classical benchmarks that provides a fair test-bed on robustness. All sequences in the dataset are labeled with 11 interference attributes, including illumination variation (IV), scale variation (SV), occlusion (OCC), deformation (DEF), motion blur (MB), fast motion (FM), in-plane rotation (IPR), out-of-plane rotation (OPR), out-of-view (OV), low resolution (LR) and background clutter (BC). 
A comparison with state-of-the-art trackers is shown in Figure~\ref{fig-otb2015} in terms of success plots of OPE. Our SiamGAT reaches a success score of $71.0\%$ that surpasses all other trackers.
An evaluation on different attributes is shown by Figure~\ref{fig-attributes}. Our tracker can better handle the challenges like deformation (DEF), out-of-plane rotation (OPR), occlusion (OCC), illumination variation (IV), in-plane rotation (IPR) and scale variation (SV), which may cause large shape and pose variations of the object.
Regarding to fast motion (FM), out-of-view (OV), low resolution (LR) which may cause extreme appearance variations, the proposed tracker obtains a lower score than the baseline SiamCAR. 
The results demonstrate that the proposed tracker can achieve robust performance against shape and pose variations.

\subsection{Evaluation on LaSOT}
To further evaluate the proposed approach on a more challenging dataset, we conduct experiments on LaSOT \cite{lasot}, which is a large-scale, high-quality, and densely annotated dataset for long-term tracking. To mitigate potential class bias, it provides the same number of sequences for each category.
The results on LaSOT are shown
in
Figure~\ref{fig-lasot}. Our SiamGAT is the second best
only 
behind
the online tracker Ocean-online \cite{Ocean} but surpasses the long-term tracker GlobalTrack \cite{GlobalTrack} by $3.6\%$ in normalized precision, $0.2\%$ in precision and $1.8\%$ in success. 
Compared with Ocean-offline \cite{Ocean} which is much more complex than SiamGAT in terms of network architecture, SiamGAT performs $2.3$ points better in normalized precision, $0.4$ in precision and $1.3$ in success.
The results indicate that the proposed tracker is competitive for long-term tracking tasks.
Moreover, up to our investigation, both attentive and target-aware properties of the proposed tracker allow more efficient online tracking without model updating. Online tracking modules can be easily integrated in future.

\section{Conclusion}

In this paper, we have presented a novel target-aware Siamese Graph Attention network, 
termed
SiamGAT,
for general object tracking. We provide theoretical and empirical evidences that how GAM establishes \emph{part-to-part} correspondence and enables each part of the search region to aggregate information from the target. Instead of using the traditional cross-correlation based information embedding method, our GAM realizes part-level information propagating between the two Siamese branches and yields a much effective information embedding map. By recomputing a target-aware template area that can adaptively fit with different object scales and aspect ratios, the proposed approach enables more generalizable visual tracking. Without bells and whistles, our SiamGAT outperforms state-of-the-art trackers by clear margins on multiple main-stream benchmarks including GOT-10k, UAV123, OTB-100 and LaSOT.

{\small
\bibliographystyle{ieee_fullname}
\bibliography{egbib}
}
\typeout{get arXiv to do 4 passes: Label(s) may have changed. Rerun}
\end{document}